\documentclass{ecai} 
\usepackage{latexsym}
\usepackage{amssymb}
\usepackage{amsmath}
\usepackage{amsthm}
\usepackage{booktabs}
\usepackage{enumitem}
\usepackage{graphicx}
\usepackage{color}

\newcommand{\BibTeX}{B\kern-.05em{\sc i\kern-.025em b}\kern-.08em\TeX}

\usepackage{graphicx}
\graphicspath{ {./images/} }

\usepackage{array} % for 'w' column type
\usepackage{calc}

\usepackage{float}

\newlength\mylen

\begin{document}

%%%%%%%%%%%%%%%%%%%%%%%%%%%%%%%%%%%%%%%%%%%%%%%%%%%%%%%%%%%%%%%%%%%%%%%%

\begin{frontmatter}

%%% Use this command to specify your submission number.
%%% In doubleblind mode, it will be printed on the first page.

\paperid{2563} 

%%% Use this command to specify the title of your paper.

\title{Domain-specific long text classification from sparse relevant information}

%%% Use this combinations of commands to specify all authors of your 
%%% paper. Use \fnms{} and \snm{} to indicate everyone's first names 
%%% and surname. This will help the publisher with indexing the 
%%% proceedings. Please use a reasonable approximation in case your 
%%% name does not neatly split into "first names" and "surname".
%%% Specifying your ORCID digital identifier is optional. 
%%% Use the \thanks{} command to indicate one or more corresponding 
%%% authors and their email address(es). If so desired, you can specify
%%% author contributions using the \footnote{} command.

\author[A]{\fnms{Célia}~\snm{D'Cruz}\thanks{Corresponding Author. Email: celia.dcruz@inria.fr}}
\author[B]{\fnms{Jean-Marc}~\snm{Bereder}\thanks{Corresponding Author. Email: bereder.jm@gmail.com}} 
\author[A]{\fnms{Frédéric}~\snm{Precioso}\thanks{Corresponding Author. Email: frederic.precioso@inria.fr}}
\author[A]{\fnms{Michel}~\snm{Riveill}\thanks{Corresponding Author. Email: michel.riveill@inria.fr}}

\address[A]{Université Côte d’Azur, CNRS, Inria, I3S, France}
\address[B]{CHU de Nice, France}

%%% Use this environment to include an abstract of your paper.

\begin{abstract}
Large Language Models have undoubtedly revolutionized the Natural Language Processing field, the current trend being to promote one-model-for-all tasks (sentiment analysis, translation, etc.). However, the statistical mechanisms at work in the larger language models struggle to exploit the relevant information when it is very sparse, when it is a weak signal. This is the case, for example, for the classification of long domain-specific documents, when the relevance relies on a single relevant word or on very few relevant words from technical jargon. In the medical domain, it is essential to determine whether a given report contains critical information about a patient's condition. This critical information is often based on one or few specific isolated terms. In this paper, we propose a hierarchical model which exploits a short list of potential target terms to retrieve candidate sentences and represent them into the contextualized embedding of the target term(s) they contain. A pooling of the term(s) embedding(s) entails the document representation to be classified. We evaluate our model on one public medical document benchmark in English and on one private French medical dataset. We show that our narrower hierarchical model is better than larger language models for retrieving relevant long documents in a domain-specific context.
\end{abstract}

\end{frontmatter}

%%%%%%%%%%%%%%%%%%%%%%%%%%%%%%%%%%%%%%%%%%%%%%%%%%%%%%%%%%%%%%%%%%%%%%%%

\section{Motivations and related works}
With the impact of Large Language Models (LLMs) over a wider community than the Natural Language Processing (NLP) one, even beyond the boundaries of the scientific community, there is a tendency to explore the full capabilities of these new models in particular for every NLP task. It seems that "bigger is always better" and "one-model-for-all-tasks" (e.g. the foundation models) are the new trendy mottos in NLP.
\paragraph{\textit{Domain-specific and technical jargon issues}}
Despite some remarkable successes of LLMs, when investigating the potential of these approaches for specific tasks, a very recent work~\cite{sheik2024neural} has shown that smaller language models are more appropriate in a domain-specific context, especially when high accuracy is expected because we are considering a high-stakes domain, such as law in their paper or medicine in our work. In their work, the authors consider more specifically several algorithms based on BERT backbone~\cite{devlin_bert_2019}, RoBERTa~\cite{liu_roberta_2019}, DistilBERT~\cite{sanh2019distilbert}. The authors explain the impact of considering a technical jargon, called \textit{“Legalese language"} in Law domain, where words which are similar in general English cannot be considered as similar by the model because they are two distinct categories (of crime) in legal texts, as for instance theft and fraud~\cite{sheik2024neural}. Such configurations are obvious challenges for very large language models pre-trained on a huge amount of data in general English.

We face the exact same problems with the technical jargon present in the medical reports. Patient medical reports contain a significant amount of information on diverse health characteristics in structured text such as key-value pairs (e.g. lab results) or in unstructured free text (e.g medical history). However, specific information contained in free text can be hard to retrieve. Reports can often be several pages long, and similar terms in general English can appear in different places of the same report confusing the language model on the final relevance of the report. Examples can be found in the literature: although anterior and ventral are similar in meaning, anterior is used to indicate spatial relationships of body parts to each other throughout the body, but ventral is most often used to indicate a relationship to the anterior abdominal wall; the use of upper rather than superior has evolved from common usage rather than following a strict anatomical designation. For instance, upper lip and upper eyelid are proper clinical designations rather than superior lip or superior eyelid, but superior pole of the kidney is the proper terminology for the “upper” or “top” of the kidney.

We also face the dual phenomenon that the wording in free text may widely differ from one report to another, both in the linguistic structure and in the medical terms. Different medical terms, as well as Latin form, or medical abbreviations (sometimes even codes), can be used to describe similar patient's condition~\cite{Nickele014129, McCafferye008094}: asthenia or weakness, arthralgia or joint pain, epistaxis or nosebleed, etc.

We do not even mention typos that can transform \textit{ileum} (the last section of the small intestine between the jejunum and the colon) into \textit{ilium} (a part of the pelvic bone), or \textit{peroneal} (pertaining to the lateral aspect of the leg) into \textit{perineal} (pertaining to the pelvic floor).

All of these issues on a rigorous use of a precise medical terminology and their consequences on health care are well documented and taken seriously in training programs in medicine~\footnote{European Medicine Agency, Medical terms simplifier, https://www.ema.europa.eu/en/documents/other/ema-medical-terms-simplifier\_en.pdf}. 

This context explains why our first attempts to investigate recent advances in information retrieval based on language models have not been conclusive~\cite{karpukhin-etal-2020-dense, qu2021rocketqa}. Still, the methods proposed in these papers are based on variants of BERT backbone (the aforementioned ones, or ALBERT~\cite{DBLP:conf/iclr/LanCGGSS20}, ERNIE \cite{DBLP:conf/acl/ZhangHLJSL19}), as for the previously cited papers.

\paragraph{\textit{Long text classification issues}}
It took only a few years for Transformer-based models \cite{NIPS2017_3f5ee243} to become state-of-the art for text encoding, drastically improving text classification performances. However, when focusing on the aforementioned encoders most of them are limited to 512 tokens as the maximum input sequence length, limiting the amount of text that can be taken into account. These text embedding methods have been mostly used for short text classification, while medical reports can be several pages long.

For classifying long texts, several strategies have been proposed. The most basic strategies involve indiscriminately truncating long documents, thus completely discarding large text parts that can contain relevant information. For instance, taking into account only the beginning, or the end, or a random part of a medical report.

Other models involve selecting key sentences from documents such as CogLTX \cite{cogltx_2020} and Bert+TextRank \cite{efficient_classif_2022}.

Other approaches adapt existing pre-trained transformer models by using hierarchical strategies \cite{yang_hierarchical_2016, gao_hierarchical_2018, pappagari_hierarchical_2019, dai_revisiting_2022}, dividing the whole text into small parts that are iteratively fed to the transformer model.

Another strategy involves the use of long-range Transformers \cite{tay_long_2021, tay_efficient_2023} that employ various strategies to reduce quadratic self-attention complexity, thus enabling them to have a significantly higher limit in the number of tokens that can be processed (e.g. Longformer \cite{beltagy_longformer_2020}, Reformer \cite{kitaev_reformer_2020}, BigBird \cite{zaheer_big_2021}). However, most of these developments have been primarily made for English.

Finally, the most recent approaches that have been proposed are based on extracting embedding layers from decoder architectures such as Llama 2~\cite{llama2}.

However, when the objective is to focus on a specific critical detail, as opposed to topic classification, relevant information represents only a small portion of all the information present in the reports. The relevant information may appear a few times in a report, or even only once. Additionally, the available training corpus size can be quite limited. In this context, most of the existing models above do not demonstrate higher performances for our peculiar classification while requesting significantly bigger computation resources.
\paragraph{\textit{Privacy and computation resources concerns}}
In \cite{schick2021s}, Schick and Schütze show that small language models (ALBERT, RoBERTa) can be few-shot learners as larger models have shown. The authors mention that they achieve similar performances as larger LLMs with \textit{``language models that are much “greener” in that their parameter count is several orders of magnitude smaller''}. In \cite{hsieh2023distilling}, Hsieh et al. emphasizes how large LLMs are \textit{"challenging to deploy in real world applications due to their sheer size"} and that their \textit{"computational requirements are far beyond affordable for most product teams, especially for applications that require low latency performance"}.

This concern is particularly relevant in our case since working on medical personal data prevents from outsourcing computations using cloud solutions, and almost no health center hosts computation resources is able to run very large language models on a regular basis. Designing a lighter model is therefore one of our objectives, and to do this, relying on one of the algorithms in the BERT family is a reasonable choice.\\

In this paper, we present our deep learning approach designed to classify medical reports, documents ranging from one to several pages in length, focusing on critical but sparse specific information, swamped with noisy and confusing content. 
\paragraph{\textit{We propose three main contributions:}}
\begin{itemize}
    \item A filtering phase to reduce the representativeness gap between relevant and irrelevant content based on a short list of target terms (up to 30), from the semantic field of the information to be retrieved. For reproducibility sake, we provide alternatives to generate this short list from publically available external resources. We show that our deep model is not much impacted by the size of an automatically generated list even if it contains extra useless terms.
    \item A hierarchical deep model based, first, on the BERT embedding of the potentially relevant target terms contained in the filtered sentences, then on the attention weight average of these term embeddings to result in the document embedding that is finally classified.
    \item An extensive evaluation and ablation study conducted against all the state-of the-art approaches (other hierarchical deep models, non-hierarchical models, long-range transformers), showing also the robustness of our model with respect to the quality of the short list of target terms used during the filtering phase, and demonstrating the overall performance of our solution. 
\end{itemize}

In this paper we detail both the private French corpus and the public English benchmark used for evaluation (Sec.~\ref{sec:datasets}), describe our approach (Sec.\ref{sec:approach}), and test the effectiveness of our approach against reference approaches on the two benchmarks (Sec.~\ref{sec:results}).
%%%%%%%%%%%%%%%%%%%%%%%%%%%%%%%%%%%%%%%%%%%%%%%%%%%%%%%%%%%%%%%%%%%%%%%%

\section{Datasets}
\label{sec:datasets}

\subsection{Public English dataset for smoking status}
We use the publicly available English dataset "2006 -Smoking" \cite{uzuner_identifying_2008}, where the classification task is to identify patient smoking status from medical discharge records. This dataset was released as part of the National NLP Clinical Challenges (n2c2, previously i2b2)
\footnote{Website for the National NLP Clinical Challenges (n2c2) \url{https://n2c2.dbmi.hms.harvard.edu/}}.

The dataset consists of 502 de-identified
discharge summaries annotated by pulmonologists to classify patient reports into five categories: \textit{Past Smoker} (patient is a former smoker who has not smoked for at least one year), \textit{Current Smoker} (patient is currently a smoker or was a smoker within the past year), \textit{Smoker} (a Current or a Past Smoker but the report does not provide enough information to classify as either), \textit{Non-Smoker} (patient has never been a smoker), \textit{Unknown} (the report does not mention anything about smoking).
There is an average of 874 words per report, with a maximum of 3292.

Some reports do not give any indication on the patient smoking status. In other reports, while this information is provided, it is mentioned once or at very few places, therefore representing only a tiny fraction of a report. Moreover, the wording indicating the patient status varies from one report to another. For instance, both sentences "She is a former smoker" and "Nicotine abuse , quit in the 80s" indicates that the patient is a past smoker.

The provided dataset contains 104 reports as testing data (11 \textit{Past Smoker}, 11 \textit{Current Smoker}, 3 \textit{Smoker}, 16 \textit{Non-Smoker}, 63 \textit{Unknown}) and 398 reports as training data (36 \textit{Past Smoker}, 35 \textit{Current Smoker}, 9 \textit{Smoker}, 66 \textit{Non-Smoker}, 252 \textit{Unknown}).

\subsection{French dataset from colorectal cancer patients}
The French dataset\footnote{This dataset is private and has been created under an agreement between Nice University Hospital and Université Côte d’Azur.} consists of 198 de-identified French medical reports from colon cancer patients obtained from Nice University Hospital. All identifying information - including patients' and doctors' names, birthdays, ID numbers, locations - has been removed. The anonymity was verified through the use of regex rules and each report was carefully read multiple times by medical experts. The medical records cover the period from the beginning of 2017 to the end of 2021. They range from one page to a few pages long (572 words on average per report, maximum 2115) with a pre-determined sequence of sections. However, the text of each section is freely written. One of the main types of reports is follow-up letters with sections describing the patient’s co-morbidities, cancer history, previous treatments, biological parameters, etc. Other report types are pathological reports, surgical reports, imaging results, etc.

Those reports may contain information about the \textit{laterality} of the primary colon tumor (classified as either "left", "right", or "no information"), whether a \textit{surgery} for the cancer has already been performed (classified as either "yes" or "no / no information"), and whether a \textit{chemotherapy} has already been initiated (classified as either "yes" or "no / no information"). Some reports do not indicate any of this information, while other reports indicate several instances of this information in one or more sections.

Our dataset has been annotated by a surgical oncologist and a Deep Learning (DL) expert in two ways: the exact locations of the relevant information in each report, as well as the overall classification of the information at the report level. As an example, for the \textit{laterality} of the primary colon tumor, if a report indicates “The patient has a tumor in the left colon and a metastasis in the right lobe of the liver.”, only the word “left” is annotated as indicating the \textit{laterality} of the primary colon tumor, since the word “right” does not indicate the \textit{laterality} of the primary tumor, it refers instead to a metastasis. The whole report is classified as “left” for the \textit{laterality}.

Out of 198 reports, the document label distribution for \textit{laterality} is 67 (right), 81 (left), and 50 (no). The distribution for \textit{surgery} is 44 (no) and 154 (yes) and for \textit{chemotherapy} it is 145 (no) and 53 (yes).

Although the private French medical dataset cannot be shared, we still present the results as it is interesting in the sense that the list of relevant keywords was manually defined by medical experts, and that the classification task \textit{laterality} is particularly challenging since "left" and "right" are target terms but are also everywhere in reports.

\section{Approach}
\label{sec:approach}

\subsection{Overview}

For contexts where the relevant information appears quite sparsely in the documents, models that indiscriminately take into account the whole text will likely struggle to focus on the few relevant parts to accurately classify the documents. The idea of our approach is to only keep the potentially relevant parts of a document while discarding most of the remaining probably irrelevant parts, therefore enabling our transformer-based deep learning model to focus only on a fraction of a document.

\begin{figure*}
    \includegraphics[width=\textwidth] {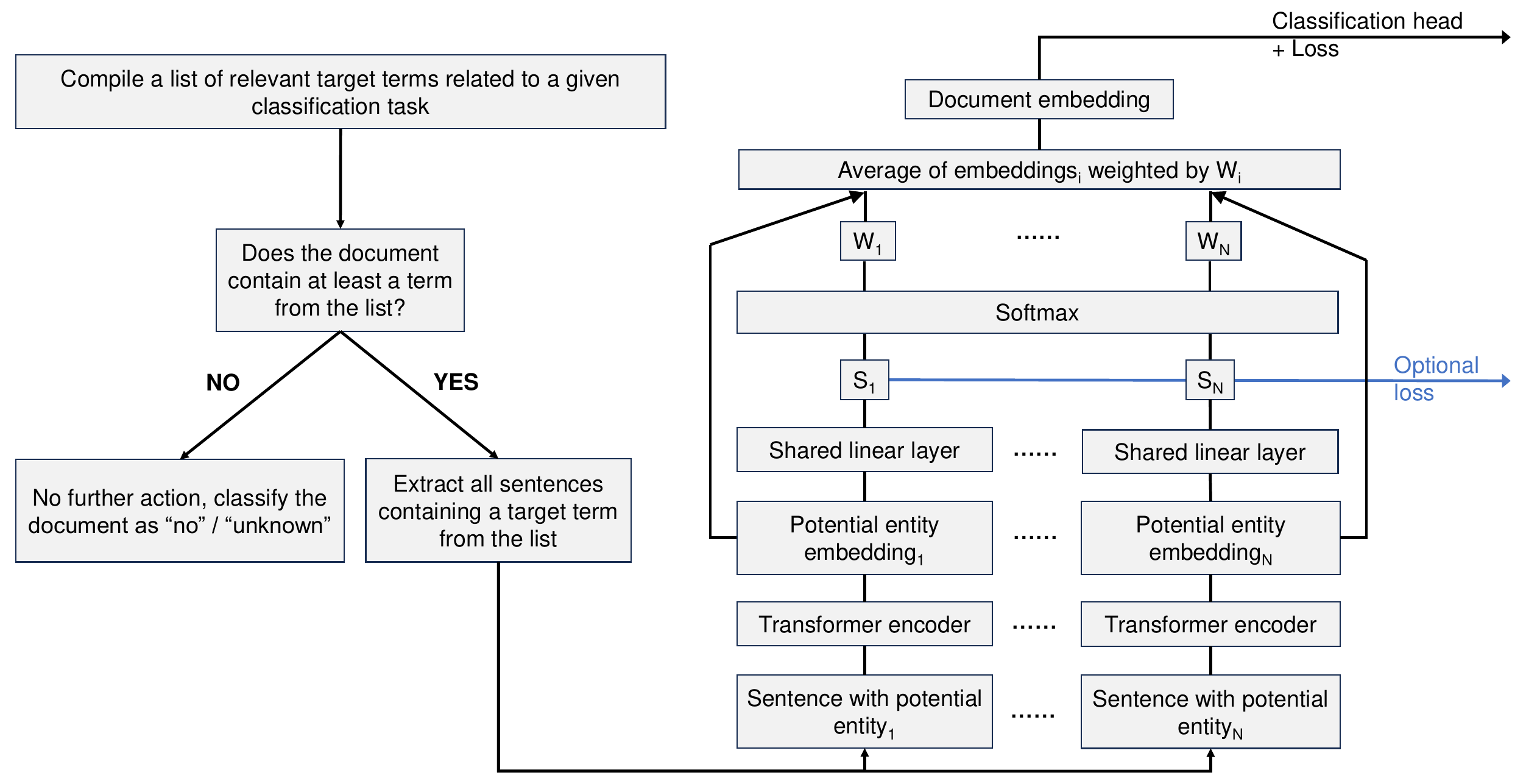}
    \caption{Architecture of our approach}
    \label{fig:architecture}
% \vspace{-0.5cm}
\end{figure*}

The relevant parts of a document are retrieved by building a list of target terms potentially relevant to the classification task, and then by extracting all the sentences containing one or more of those target terms.
Figure \ref{fig:architecture} illustrates our approach.

% https://tex.stackexchange.com/questions/30985/displaying-a-wide-figure-in-a-two-column-document

\subsection{Relevant target term list}
We compile a list of target terms relevant to each classification task.
It should be noted that those target terms can be single words or composed of several words.

\subsubsection{English dataset relevant terms}
The list of relevant terms aims at identifying the patient smoking status in the English dataset. We test two different approaches. In the first approach, the relevant terms are automatically extracted from the website relatedwords.io\footnote{\url{https://relatedwords.io}} that gathers, from a single source target term, a list of related terms sorted by relevance. In our case, since we want to classify the smoking status of patients (Current Smoker, Past Smoker, Smoker, Non-Smoker, Unknown), we choose the source target term "smoker". In the second approach, we do not use any external resource; we use the X-Class \cite{x_class} keyword expansion method with the source target term "smoker" to retrieve relevant terms related to "smoker" within the dataset from the training document themselves.

We remove English stop words from the aforementioned list of relevant target terms using the Python library stopwordsiso \footnote{\url{https://pypi.org/project/stopwordsiso/}} that removes terms such as "not", "non", "very".
Therefore, the final list contains the source target term "smoker" and its related terms from the relatedwords.io website minus the stopwords.

This final target term list is ranked by the relevance of the terms to the words "smoker" in both relatedwords.io website and X-Class methods. However, we can vary the list size by removing the terms ranked as least relevant in order to study the sensitivity of our model to the target term list, as we did in our experiments.

\subsubsection{French dataset relevant terms}

The list of relevant target terms for each classification task -- \textit{laterality} (of the primary colon tumor), \textit{surgery} (performed), and \textit{chemotherapy} (initiated) -- to identify relevant information from the French dataset has been provided by the medical experts from Nice University Hospital, who considered that list as more relevant than using a French version of relatedwords.io\footnote{For instance : https://www.rimessolides.com/motscles.aspx?m=chirurgie}. The three lists provided by the experts range from approximately 10 to 20 terms. Those lists are much smaller than the ones extracted from relatedwords.io because those three lists have been carefully tailored to each classification task, which are quite specific. Those lists provided by the experts are not ranked by relevance. We simulate variable list sizes by manually adding terms that are of little relevance to the classification task (namely "augmented vocabulary"), or manually deleting some relevant terms from those lists (namely "reduced vocabulary").

\subsection{Filtering}

Once the list of relevant target terms has been established, we search in the documents for all instances of those target terms.

\textbf{\textit{Case A:}} if a document does not contain any target term, it is categorized as "no / no information" regarding laterality, surgery, or chemotherapy, or "unknown" regarding the patient smoking status. No further action is taken.

\textbf{\textit{Case B:}} if target terms are present, we extract all sentences from the documents that include these target terms. These sentences provide a context. These target terms may be relevant, because they actually provide explicit information on the label of the document for one of the target classes (e.g. the laterality, the occurrence of a surgery, or the initiation of chemotherapy), but they may also be irrelevant (“past shoulder surgery after road accident” does not indicate a surgery for cancer, “the use of a chemotherapy treatment will later be discussed” does not indicate that a chemotherapy has already been initiated). The target term "smoker" can be used in sentences such as "she is a former smoker", "she is not a smoker", "current smoker", which does not lead to the same smoking status label. From now on, we call \textit{entities} the target terms in the context of their surrounding sentence.

\subsection{Model architecture}
\subsubsection{Main architecture}

The filtering done in the previous step retained the sentences containing the relevant vocabulary. We obtain a contextualized embedding of each target term by passing the sentence to which it belongs to a transformer encoder and selecting specifically the embedding associated to the target term.

Some entities may actually be irrelevant to the classification task, while others are highly relevant. The model calculates an attention weight $W_i$ for each target term to give more weights to relevant entities. This is done by passing each target term embedding to a shared linear layer that outputs a score $S_i$ and normalizing all the scores with a softmax layer.

The document embedding is the product sum of each entity embedding and their attention weight. This vector is then used in the classification head that includes a linear layer followed by a softmax layer with cross-entropy loss function.

\subsubsection{Model extension}

While a first filtering phase has been performed to select potentially relevant sentences based on the presence of target terms, it might still be challenging in certain cases to obtain very good classification results. For instance, the laterality classification task is challenging because the target terms "right" and "left" may very often appear within a report but be relevant to the primary tumor laterality only in a few sentences. Therefore, a second filtering phase can be optionally performed by submitting the sentences in the training set initially selected from the first filtering phase to the experts for further manual annotation to determine whether the terms within their sentence are indeed relevant or not. However, to reduce the need of interactions with the experts, an active learning strategy can be employed to annotate only a small fraction of the documents during this second filtering phase. 

Our initial model architecture can be extended by adding a sigmoid function after the $S_i$ scores followed by an optional binary cross entropy loss to leverage the annotation of the second optional filtering phase at training time. Table \ref{tab:ner_classification} shows the improved performance on our three document classification tasks in our French dataset that has been further annotated for the second optional filtering phase.

\begin{table*}[!ht]
\begin{center}
\renewcommand\arraystretch{1.5} % default: 1
\small % \tiny % optional
% calculate usable col. width:
\setlength\tabcolsep{40pt} % default: 6pt
\setlength\mylen{(\textwidth-16\tabcolsep-9\arrayrulewidth)/15}
\begin{tabular}{| *{7}{w{c}{\mylen}|} }
\hline
\multicolumn{1}{|c|}{} &
\multicolumn{2}{|c|}{\begin{tabular}[x]{@{}c@{}}Laterality\end{tabular}} &
\multicolumn{2}{|c|}{\begin{tabular}[x]{@{}c@{}}Surgery\end{tabular}} &
\multicolumn{2}{ c|}{\begin{tabular}[x]{@{}c@{}}Chemotherapy\end{tabular}} \\
\hline
 & Accuracy & \begin{tabular}[x]{@{}c@{}}Balanced\\Accuracy\end{tabular} & Accuracy & \begin{tabular}[x]{@{}c@{}}Balanced\\Accuracy\end{tabular} & Accuracy & \begin{tabular}[x]{@{}c@{}}Balanced\\Accuracy\end{tabular} \\
\hline
LR & 0.556 & 0.542 & 0.843 & 0.753 & \underline{0.874} & \underline{0.872} \\
\hline
HiSAN & 0.526 & 0.490 & 0.847 & 0.761 & 0.866 & 0.841 \\
\hline
Transformer over CLS & 0.561 & 0.543 & 0.866 & 0.737 & 0.851 & 0.820 \\
\hline
SAN & 0.579 & 0.544 & 0.860 & \underline{0.866} & 0.839 & 0.834 \\
\hline
Bert+TextRank & \underline{0.689} & \underline{0.662} & \underline{0.878} & 0.787 & 0.865 & 0.862 \\
\hline
Ours & \textbf{0.782} & \textbf{0.741} & \textbf{0.940} & \textbf{0.892} & \textbf{0.925} & \textbf{0.875} \\
\hline
\end{tabular}
\caption{Model comparisons on individual tasks on our French dataset}
\label{tab:french_models_results}
%\vspace{-0.4cm}
\end{center}
\end{table*}

%%%%%%%%%%%%%%%%%%%%%%%%%%%%%%%%%%%%%%%%%%%%%%%%%%%%%%%%%%%%%%%%%%%%%%%%%%%%%
\section{Results}
\label{sec:results}

\subsection{Baseline model and DL contenders}

We benchmark our approach on both our French and English datasets against Machine Learning (ML) and DL architectures designed for long text classification.

\begin{itemize}

\item \textbf{ML model}. Logistic regression (LR) where we removed beforehand stopwords and punctuations from the reports, we lemmatized and changed the text to lowercase, and applied TF-IDF.

\item \textbf{Sentence selection DL model}. Bert+TextRank \cite{efficient_classif_2022}.

\item \textbf{Flat DL model}. "SAN"-like model \cite{gao_classifying_2019} that performs classification on the document embedding obtained by simply performing a weighted average of all subword embeddings.

\item \textbf{Hierarchical DL models}. 1) "HiSAN"-like model \cite{gao_classifying_2019} with a Transformer encoder to replace their non-contextual embeddings and we split the texts into blocks of 512 tokens. 2) "Transformer over CLS" with max pooling \cite{dai_hierarchical_2022}.

\item \textbf{Long-range transformers}. 1) Clinical-LongFormer using global attention only on the first token \cite{li_clinical_bigbird_longformer_nodate}. 2) Clinical-BigBird using sparse attention \cite{li_clinical_bigbird_longformer_nodate}. 3) Llama 2 7B \cite{llama2} using the embedding of the last token. If a text exceeds 4096 tokens, only the beginning of the text up to 4096 tokens is used as input to the models.

\end{itemize}

\subsection{Experimental setup}

We use the test set provided in the National NLP Clinical Challenges (n2c2), for the English "2006-Smoking" dataset. We apply a 5-fold cross-validation procedure on the remaining training / validation set and we report the average results on the test set.
We apply a 5-fold cross-validation procedure on the French dataset where the sum of each test fold covers the entire dataset.

Except for ML and long-range transformer approaches, we use the Bio+Discharge Summary BERT \cite{alsentzer-etal-2019-publicly} pretrained model from Hugging Face library\footnote{\url{https://huggingface.co/emilyalsentzer/Bio_Discharge_Summary_BERT}} as the Transformer encoder for the English dataset, and the CamemBERT-base pretrained model \cite{martin_camembert_2020} from Hugging Face library\footnote{\url{https://huggingface.co/camembert-base}} as the Transformer encoder for our French dataset.
We use PySBD \cite{pysbd_2020} to identify sentence boundaries.

For each classification task and for each architecture, we train a separate model. Our evaluation metrics are the accuracy\footnote{\url{https://scikit-learn.org/stable/modules/generated/sklearn.metrics.accuracy_score.html}} and balanced accuracy\footnote{\url{https://scikit-learn.org/stable/modules/generated/sklearn.metrics.balanced_accuracy_score.html}}. In all DL experiments, we use the same optimizer (AdamW), effective batch size (4), learning rate (2e-5), early stop patience (5) and label smoothing (0.1).

\subsection{Experimental results}
\begin{table}[!ht]
\begin{center}
\renewcommand\arraystretch{1.5} % default: 1
\small % \tiny % optional
% calculate usable col. width:
\setlength\tabcolsep{40pt} % default: 6pt
\setlength\mylen{(\textwidth-16\tabcolsep-9\arrayrulewidth)/15}
\begin{tabular}{| *{3}{w{c}{\mylen}|} }
\hline
\multicolumn{1}{|c|}{} &
\multicolumn{2}{|c|}{Smoking status} \\
\hline
 & Accuracy & \begin{tabular}[x]{@{}c@{}}Balanced\\Accuracy\end{tabular} \\
\hline
LR & 0.596 & 0.215 \\
\hline
Bert+TextRank & 0.604 & 0.253 \\
\hline
SAN & 0.817 & 0.551 \\
\hline
Tranformer over CLS & 0.848 & 0.624 \\
\hline
HiSAN & 0.823 & 0.577 \\
\hline
Llama 2 & 0.829 & 0.560 \\
\hline
LongFormer & 0.813 & 0.599 \\
\hline
BigBird & 0.875 & 0.652 \\
\hline
\begin{tabular}[x]{@{}c@{}}Ours (N = 30)\\ relatedWords.io\end{tabular} & \textbf{0.900} & \underline{0.665} \\
\hline
\begin{tabular}[x]{@{}c@{}}Ours (N = 20)\\ X-Class\end{tabular} & \underline{0.887} & \textbf{0.668} \\
\hline
\end{tabular}
\caption{Model comparisons on the English dataset. "Ours" corresponds to our approach with the first N most relevant target terms extracted from the relatedwords.io website or X-Class method}
\label{tab:english_models_results}
% \vspace{-0.4cm} 
\end{center}
\end{table}

\subsubsection{Comparison with diverse types of approach}

We compare our approach with different types of models as illustrated in Table \ref{tab:french_models_results} on the French dataset and Table \ref{tab:english_models_results} on the English dataset. Our approach shows substantially better results than ML, flat, or other sentence selection models. Likewise, our narrower hierarchical model, which uses a fraction of the text by only selecting potentially relevant sentences, demonstrates better results than other hierarchical models where the whole text is indiscriminately fed iteratively to a Bert-like model. Long-range transformers allow the entire documents to be processed all at once, but they are computationally expensive and our lighter approach shows better performances.

\begin{table*}[!ht]
\begin{center}
\renewcommand\arraystretch{1.5} % default: 1
\small % \tiny % optional
% calculate usable col. width:
\setlength\tabcolsep{25pt} % default: 6pt
\setlength\mylen{(\textwidth-16\tabcolsep-9\arrayrulewidth)/15}
\begin{tabular}{| *{5}{w{c}{\mylen}|} }
\hline
\multicolumn{1}{|c|}{} &
\multicolumn{2}{|c|}{\begin{tabular}[x]{@{}c@{}}relatedwords.io\end{tabular}} &
\multicolumn{2}{|c|}{\begin{tabular}[x]{@{}c@{}}X-Class\end{tabular}}
\\
\hline
& Accuracy & \begin{tabular}[x]{@{}c@{}}Balanced\\Accuracy\end{tabular} & Accuracy & \begin{tabular}[x]{@{}c@{}}Balanced\\Accuracy\end{tabular} \\
\hline
Ours (N = 10) & 0.788 & 0.489 & 0.887 & 0.658 \\
\hline
Ours (N = 20) & 0.894 & 0.655 & 0.887 & 0.668 \\
\hline
Ours (N = 30) & 0.900 & 0.665 & 0.879 & 0.659 \\
\hline
Ours (N = 40) & 0.887 & 0.665 & 0.867 & 0.639 \\
\hline
Ours (N = 50) & 0.887 & 0.658 & 0.871 & 0.640 \\
\hline
\end{tabular}
\caption{Variation of the number N of most relevant target terms on the English dataset}
\label{tab:relatedwords_xclass_comparison}
%\vspace{-0.4cm}    
\end{center}
\end{table*}

\begin{table}[!ht]
\begin{center}
\renewcommand\arraystretch{1.5} % default: 1
\small % \tiny % optional
% calculate usable col. width:
\setlength\tabcolsep{25pt} % default: 6pt
\setlength\mylen{(\textwidth-16\tabcolsep-9\arrayrulewidth)/15}
\begin{tabular}{| *{3}{w{c}{\mylen}|} }
\hline
\multicolumn{1}{|c|}{} &
\multicolumn{2}{|c|}{Laterality} \\
\hline
 & Accuracy & \begin{tabular}[x]{@{}c@{}}Balanced\\Accuracy\end{tabular} \\
\hline
Ours & \textbf{0.782} & \textbf{0.741} \\
\hline
\begin{tabular}[x]{@{}c@{}}Augmented\\vocabulary\end{tabular} & 0.752 & 0.701 \\
\hline
\begin{tabular}[x]{@{}c@{}}Reduced\\vocabulary\end{tabular} & \underline{0.764} & \underline{0.723} \\
\hline
\end{tabular}
\caption{Vocabulary list variations for the laterality classification task}
\label{tab:laterality_models_results}
%\vspace{-0.4cm}
\end{center}
\end{table}

\begin{table}[!ht]
\begin{center}
\renewcommand\arraystretch{1.5} % default: 1
\small % \tiny % optional
% calculate usable col. width:
\setlength\tabcolsep{25pt} % default: 6pt
\setlength\mylen{(\textwidth-16\tabcolsep-9\arrayrulewidth)/15}
\begin{tabular}{| *{3}{w{c}{\mylen}|} }
\hline
\multicolumn{1}{|c|}{} &
\multicolumn{2}{|c|}{Surgery} \\
\hline
 & Accuracy & \begin{tabular}[x]{@{}c@{}}Balanced\\Accuracy\end{tabular} \\
\hline
Ours & \underline{0.940} & \textbf{0.892} \\
\hline
\begin{tabular}[x]{@{}c@{}}Augmented\\vocabulary\end{tabular} & 0.927 & 0.874 \\
\hline
\begin{tabular}[x]{@{}c@{}}Reduced\\vocabulary\end{tabular} & \textbf{0.942} & \underline{0.887} \\
\hline
\end{tabular}
\caption{Vocabulary list variations for the surgery classification task}
\label{tab:surgery_models_results}
%\vspace{-0.4cm}
\end{center}
\end{table}

\begin{table}[!ht]
\begin{center}
\renewcommand\arraystretch{1.5} % default: 1
\small % \tiny % optional
% calculate usable col. width:
\setlength\tabcolsep{30pt} % default: 6pt
\setlength\mylen{(\textwidth-16\tabcolsep-9\arrayrulewidth)/15}
\begin{tabular}{| *{3}{w{c}{\mylen}|} }
\hline
\multicolumn{1}{|c|}{} &
\multicolumn{2}{|c|}{Chemotherapy} \\
\hline
 & Accuracy & \begin{tabular}[x]{@{}c@{}}Balanced\\Accuracy\end{tabular} \\
\hline
Ours & \textbf{0.925} & \textbf{0.875} \\
\hline
\begin{tabular}[x]{@{}c@{}}Reduced\\vocabulary\end{tabular}  & \underline{0.921} & \underline{0.871} \\
\hline
\begin{tabular}[x]{@{}c@{}}Further Reduced\\vocabulary\end{tabular}  & 0.902 & 0.841 \\
\hline
\end{tabular}
\caption{Vocabulary list variations for the chemotherapy classification task}
\label{tab:chemotherapy_models_results}
%\vspace{-0.4cm}
\end{center}
\end{table}

\begin{table*}[!ht]
\begin{center}
\renewcommand\arraystretch{1.5} % default: 1
\small % \tiny % optional
% calculate usable col. width:
\setlength\tabcolsep{30pt} % default: 6pt
\setlength\mylen{(\textwidth-16\tabcolsep-9\arrayrulewidth)/15}
\begin{tabular}{| *{7}{w{c}{\mylen}|} }
\hline
\multicolumn{1}{|c|}{} &
\multicolumn{2}{|c|}{\begin{tabular}[x]{@{}c@{}}Laterality\end{tabular}} &
\multicolumn{2}{|c|}{\begin{tabular}[x]{@{}c@{}}Surgery\end{tabular}} &
\multicolumn{2}{ c|}{\begin{tabular}[x]{@{}c@{}}Chemotherapy\end{tabular}} \\
\hline
 & Accuracy & \begin{tabular}[x]{@{}c@{}}Balanced\\Accuracy\end{tabular} & Accuracy & \begin{tabular}[x]{@{}c@{}}Balanced\\Accuracy\end{tabular} & Accuracy & \begin{tabular}[x]{@{}c@{}}Balanced\\Accuracy\end{tabular} \\
\hline
Ours & 0.782 & 0.741 & 0.940 & 0.892 & 0.925 & 0.875 \\
\hline
Ours + 2nd loss & \textbf{0.904} & \textbf{0.891} & \textbf{0.979} & \textbf{0.970} & \textbf{0.940} & \textbf{0.901} \\
\hline
\end{tabular}
\caption{Document classification results of our model with the optional second loss function}
\label{tab:ner_classification}
%\vspace{-0.4cm}    
\end{center}
\end{table*}

\subsubsection{Robustness of the vocabulary list}

Both the French and English datasets show that our approach is not very sensitive to noise in the target term list, namely the vocabulary list. The inclusion of additional unnecessary terms or the omission of non-critical terms does not have a major impact on the results.

Table \ref{tab:relatedwords_xclass_comparison} shows, on the English datatset, the impact of the variation of the number N of most relevant target terms generated either by the relatedwords.io website or the X-Class keyword expansion method.

With the relatedwords.io approach on the English dataset, the first N = 10 target terms does not include some critical terms such as "smoking", hence resulting in poor results. The first N = 20 target terms does include "smoking" as well as "nonsmoker", therefore substantially improving the results. The first N = 30 target terms includes additionally relevant terms and little noise, giving the best results. Increasing N only slightly decreases the performances as noise increases with less relevant terms.

With the X-Class target term expansion method on the English dataset, the first N = 10 terms already includes critical terms such as "smoking", "tobacco", "cigarette", therefore already showing very good results. Using N = 20 shows similar results, and using additionally less relevant terms only slightly decreases the performance.

Tables \ref{tab:laterality_models_results}, \ref{tab:surgery_models_results} and \ref{tab:chemotherapy_models_results} show the impact of the variation of the vocabulary lists on the French dataset. "Reduced vocabulary" lists only include the most relevant target terms that are present in all the relevant documents, "Ours" lists include additionally relevant terms, while "augmented vocabulary" lists include additional terms that are of little relevance. The supplementary material details the vocabulary lists used in the French dataset. For instance, in the case of the laterality classification task, our "reduced vocabulary" list only contains the "left" and "right" target terms as they are the most present in the reports. "Ours" list adds the target terms related to anatomical parts of the colon that indicates the laterality (“caecum”, “sigmoid”, etc.). Our "augmented vocabulary" list contains a few additional terms that are related to the colon but are not relevant in the reports we have, such as "transverse" (middle part of the colon, neither left or right).

Furthermore, our approach is not very sensitive to the method used to find the target terms: interviewing experts from the domain, using the X-Class keyword expansion method, or using the relatedwords.io website yield strong results as long as the vocabulary lists contain enough critically relevant terms.

\subsubsection{Impact of the sparsity of the relevant information}

Models that indiscriminately consider the whole text may struggle to classify documents when the relevant information appears sparsely and the rest of the text offers very little hints. The impact the degree of sparsity is best illustrated by Table \ref{tab:french_models_results} on the French dataset. For the laterality task, the structure or type of reports (e.g. follow-up letters, imaging reports, etc.) provide very limited clues on the laterality of the colon tumor, and the information of the colon tumor laterality appears at most only in a few sentences in reports that can be a few pages long, making it challenging for other models to classify the documents, whereas our approach shows a large improvement. However, for the chemotherapy and surgery tasks, certain types of reports offer some clues. For instance, pathological reports are often related to a surgery and some types of follow-up letters are usually related to a chemotherapy. In this case, our model still shows better performance, but the improvement over other models is less substantial.

\subsubsection{Optional second loss function}
The colon tumor laterality classification task is particularly challenging since "right" and "left" terms appear everywhere in the reports and are often not relevant (e.g. “right” or “left” in a text may refers to the colon tumor laterality which is relevant, but also to the side of a lung metastasis which is irrelevant). To cope with this issue, we have thus evaluated the impact of a further refinement process involving a second targeted annotation of the sentences selected through the first relevant term list filtering, in the training set only. We have used this refined annotation with an additional loss function to help the model training on more precisely “relevant” / “non relevant” terms and sentences from the text. Since this require extra human operations, we only evaluated this process on the French dataset.

Training the model with both losses (the initial classification one, and the refinement filtering one where all filtered sentences have been further annotated in the training set), our approach shows, in Table \ref{tab:ner_classification}, significant improvements for all three classification tasks compared to using only the classification loss function, especially for the \textit{laterality} classification task as expected but still significantly also for \textit{surgery} and \textit{chemotherapy}.

An interesting future work would be designing an active learning strategy to annotate only a small subset of filtered sentences, requesting little work from domain experts for a significant improvement.

\section{Conclusion}
In this paper, we present an approach to identify if long documents are useful for experts based on relevant information which appears sparsely in those documents, as it is the case for instance in medical reports. We tackle this task considering it as a binary classification task: the document analyzed contains or not the relevant information.

Traditional ML and DL models designed for long text classification struggle in this context. Our model is based on hierarchical representations combining token-level embeddings with a sentence-level representation into a document embedding allowing to classify precisely the document as relevant or irrelevant. This specific structure combining different embedding levels of the text outperforms long range transformers as well as hierarchical deep models that have been proposed in the literature for long document classification. 

Text embedding in our model is based on BERT variants, nowadays so called small language models, and outperforms more recent LLM-based embeddings, hence confirming recent recurrent results in similar contexts that larger LLMs are not always better.    

We even propose a human (expert) in the loop mechanism for contexts where the vocabulary is confusing based on too general and broad target terms, improving the performances even more. Our approach outperforms SOTA on both a very specific French dataset and a more classic public English benchmark, and can easily be extended to other classification tasks and datasets.

In future work, we plan to investigate the extra load of refining annotation of selected sentences through an active learning strategy.

\section{Ethical considerations}

Our work holds the potential to prevent healthcare professionals from unintentionally overlooking critical details in patients’ files, therefore potentially enhancing clinical decision-making. However, our model is not flawless, and the mistakes it makes could conversely negatively impact clinical decisions. Hence, our work should complement, not replace, expert judgment. This research should be seen as a step towards facilitated information extraction while emphasizing the irreplaceability of human expertise in medical contexts.

Moreover, our private French dataset is de-identified to ensure patient confidentiality and complies with General Data Protection Regulation and the strict rules of processing medical data in France. The raw data are not processed out of the hospital, but concerns over data privacy and potential misuse could still be raised. We are aware of those concerns and all this work is conducted under a strict privacy and confidentiality procedure of the hospital.

%%%%%%%%%%%%%%%%%%%%%%%%%%%%%%%%%%%%%%%%%%%%%%%%%%%%%%%%%%%%%%%%%%%%%%%%

%%% Use this command to include your bibliography file.

\bibliography{m2563}

\newpage
\appendix

\section*{Supplementary Material: list of target terms used in the French dataset}

\label{sec:appendix}

\label{sec:appendix_keywords}

\begin{itemize}

    \item Target term list used for the laterality classification task:
    
    \begin{itemize}
    
        \item The "reduced vocabulary" list only contains:
        
        \begin{itemize}
            \item left
            \item right
        \end{itemize}
        
        \item "Ours" contains the aforementioned vocabulary and the following additional terms:
        
        \begin{itemize}
            \item words derived from "caecum": caecum, cecal
            \item words derived from "sigmoid": sigmoid, rectosigmoid, sigmoidectomy, etc.
            \item iliac
            \item bauhin
            \item ascending
            \item descending
            \item words derived from "ileo": ileo, ileocecal, ileocolic, etc.
            \item cm from the anal margin
        \end{itemize}
        
        \item The "augmented vocabulary" list contains the aforementioned vocabulary and the following additional terms:
        
        \begin{itemize}
            \item transverse
            \item splenic
            \item hepatic
            \item pelvic
            \item lumbar
        \end{itemize}
    \end{itemize}
    
    \item Target term list used for the surgery classification task:
    
    \begin{itemize}
        \item The "reduced vocabulary" list only contains:
        
        \begin{itemize}
            \item words derived from "surgery": surgery, surgical
            \item words derived from "operation": operation, operated, operating
            \item words ending with "ectomy": colectomy, sigmoidectomy, hepatectomy, metastasectomy, etc.
        \end{itemize}

        \item "Ours" contains the aforementioned vocabulary and the following additional terms:

        \begin{itemize}
            \item words derived from "resection" : "resection", "resected"
            \item exenteration
            \item cytoreduction
        \end{itemize}

        \item The "augmented vocabulary" list contains the aforementioned vocabulary and the following additional terms:

        \begin{itemize}
            \item words ending with "otomy": laparotomy, etc
            \item words ending with "ostomy": colostomy, ileostomy, etc.
            \item excision
            \item ablation
            \item exeresis
        \end{itemize}
        
    \end{itemize}
    
    \item Target term list used for the chemotherapy classification task:
    
    \begin{itemize}
        \item The "further reduced vocabulary" list only contains:
        
        \begin{itemize}
            \item chemotherapy
        \end{itemize}

        \item The "further reduced vocabulary" list contains the aforementioned vocabulary and the following additional terms:

        \begin{itemize}
            \item folfox
            \item folfiri
            \item folfirinox
            \item folfoxiri
        \end{itemize}

        \item "Ours" contains the aforementioned vocabulary and the following additional terms:

        \begin{itemize}
            \item oxaliplatin
            \item cetuximab
            \item erbitux
            \item bevacizumab
            \item xelox
            \item capox
            \item capecitabine
            \item 5-FU
        \end{itemize}
        
    \end{itemize}
    
\end{itemize}

\end{document}